\pdfoutput=1

\documentclass[11pt]{article}

\usepackage[final]{acl}

\usepackage{times}
\usepackage{latexsym}
\usepackage{enumitem}
\usepackage[T1]{fontenc}

\usepackage[utf8]{inputenc}

\usepackage{microtype}

\usepackage{inconsolata}

\usepackage{graphicx}

\usepackage{xcolor}
\usepackage{soul}
\usepackage{booktabs}
\usepackage{latexsym}
\usepackage{amsmath}
\DeclareMathOperator*{\argmax}{arg\,max}

\usepackage{amssymb}
\usepackage{multirow}

\definecolor{green}{HTML}{34a853}
\definecolor{lightgreen}{HTML}{d9ead3} 
\definecolor{seagreen}{HTML}{3CB371} 
\definecolor{darkgreen2}{HTML}{38761d}
\definecolor{lightgreen2}{HTML}{b6d7a8}

\definecolor{redberry}{HTML}{cc4125} 
\definecolor{lightredberry}{HTML}{cc4125} 
\definecolor{lightred}{HTML}{d9ead3}
\definecolor{darkpurple1}{HTML}{674ea7}

\definecolor{purple}{HTML}{9900ff} 
\definecolor{lightpurple}{HTML}{b4a7d6} 
\definecolor{lightpurple1}{HTML}{8e7cc3}

\definecolor{gray}{HTML}{cccccc}
\definecolor{lightgray1}{HTML}{d9d9d9}
\definecolor{lightgray2}{HTML}{efefef}
\definecolor{darkgray4}{HTML}{434343}

\definecolor{blue}{HTML}{4285f4} 
\definecolor{darkblue}{HTML}{0b5394} 
\definecolor{lightblue}{HTML}{9fc5e8} 
\definecolor{lightblue3}{HTML}{cfe2f3} 
\definecolor{lightcornflowerblue2}{HTML}{a4c2f4} 
\definecolor{lightcornflowerblue3}{HTML}{c9daf8} 
\definecolor{darkcornflowerblue3}{HTML}{1c4587} 

\definecolor{orange}{HTML}{ff9900} 
\definecolor{lightorange2}{HTML}{f9cb9c} 
\definecolor{lightorange3}{HTML}{fce5cd} 
\definecolor{darkorange}{HTML}{FF8C00} 
\definecolor{darkorange1}{HTML}{e69138} 

\definecolor{lightyellow2}{HTML}{ffe599}
\definecolor{lightyellow3}{HTML}{fff2cc}

\newcommand{\hlc}[2][yellow]{{%
    \colorlet{foo}{#1}%
    \sethlcolor{foo}\hl{#2}}%
}
\usepackage{url}
\usepackage{hyperref}
\usepackage{xurl}
\usepackage{inconsolata}
\usepackage{graphicx}
\usepackage{caption}
\usepackage{subcaption}
\usepackage{pgfplots}
\usepackage{tikz}
\usepackage[nameinlink]{cleveref}
\usepgfplotslibrary{groupplots}

\usepackage{tikz}
\usepgfplotslibrary{fillbetween}

\usepackage{microtype}
\usepackage{amsmath}
\usepackage{amssymb}
\usepackage{graphicx}
\usepackage{subcaption}
\usetikzlibrary{patterns}
\usepackage{filecontents}
\usepackage[noend]{algorithmic}
\usepackage{algorithm}
\usepackage{csvsimple}
\usepackage{xspace}
\usepackage{mathtools}

\DeclareUnicodeCharacter{2212}{−}
\usepgfplotslibrary{groupplots,dateplot}
\usetikzlibrary{patterns,shapes.arrows}
\pgfplotsset{compat=newest}

\title{Open-World Evaluation for Retrieving Diverse Perspectives}

\ifx\nocomment\undefined
\NewDocumentCommand{\ec}
{ mO{} }{\textcolor{cyan}{\textsuperscript{\textit{Eunsol}}\textsf{\textbf{\small[#1]}}}}
\NewDocumentCommand{\hc}
{ mO{} }{\textcolor{blue}{\textsuperscript{\textit{Hungting}}\textsf{\textbf{\small[#1]}}}}
\NewDocumentCommand{\td}
{ mO{} }{\textcolor{lightred}{\textsuperscript{\textit{TODO}}\textsf{\textbf{\small[#1]}}}}
\newcommand{\bench}{\textsc{BeRDS}}

\author{
Hung-Ting Chen \and Eunsol Choi \\
Department of Computer Science \\
New York University \\
{\texttt{\{hung-ting.chen, eunsol\}@nyu.edu}}
}

\begin{document}
\maketitle
\begin{abstract}
We study retrieving a set of documents that covers various perspectives on a complex and contentious question (e.g., will ChatGPT do more harm than good?). We curate a \textbf{Be}nchmark for \textbf{R}etrieval \textbf{D}iversity for \textbf{S}ubjective questions (\textbf{\bench}), where each example consists of a question and diverse perspectives associated with the question, sourced from survey questions and debate websites. On this data, retrievers paired with a corpus are evaluated to surface a document set that contains diverse perspectives. Our framing diverges from most retrieval tasks in that document relevancy cannot be decided by simple string matches to references. Instead, we build a language model-based automatic evaluator that decides whether each retrieved document contains a perspective. This allows us to evaluate the performance of three different types of corpus (Wikipedia, web snapshot, and corpus constructed on the fly with retrieved pages from the search engine) paired with retrievers. Retrieving diverse documents remains challenging, with the outputs from existing retrievers covering all perspectives on only 40\% of the examples. We further study the effectiveness of query expansion and diversity-focused reranking approaches and analyze retriever sycophancy. 
\end{abstract}
\section{Introduction} \label{sec:intro}

\begin{figure*}[t!]
    \centering
    \includegraphics[trim={0cm 0cm 0 0cm},clip,width=0.93\textwidth]{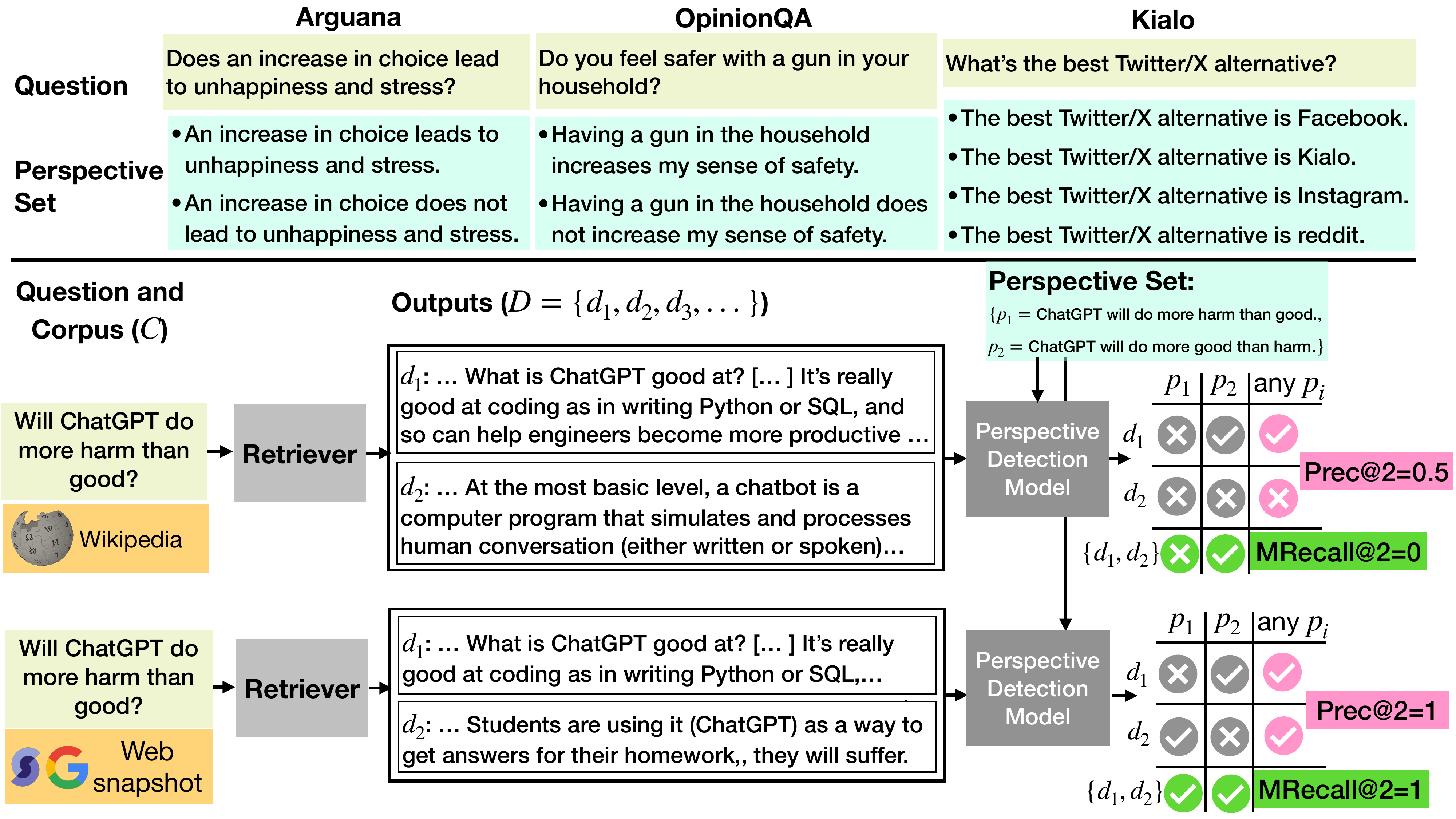}
    \vspace{-0.4em}
    \caption{Overview of our benchmark and task. The upper part shows example instances, where each instance consists of a question and a perspective set. The lower part illustrates the task. The retriever will return a set of documents, and we evaluate whether the retrieved set (in this example, consisting of two documents) contains multiple answers or perspectives. MRecall measures the coverage of answers and perspectives, and precision measures whether each document contains any perspectives. A separate module (denoted as ``Perspective Detection'' (Section~\ref{ssec:perspective-detection})) is developed to determine whether a document contains a certain perspective. We experiment with retrieving from the Web in addition to Wikipedia, which provides insufficient information for the task. } 
    \label{fig:intro}
\vspace{-1.2em}
\end{figure*}

Given a complex and contentious question~\cite{xu2024debateqa}, such as ``Will ChatGPT do more harm than good?'', a retrieval system should be able to surface diverse opinions in their top retrieval outputs. However, existing information retrieval (IR) tasks and systems mostly optimize for relevance to the question, ignoring diversity. 
A diverse IR system could be useful in two ways: (1) by surfacing diverse documents, which can be useful to users directly~\cite{chen-etal-2022-design}, and (2) by improving retrieval-augmented language models (RALMs). Prompting large language models (LLMs) to generate an answer that encompasses diverse perspectives on its own is challenging~\cite{sorensen2024roadmap,hayati2023far}, and retrieval-augmentation~\cite{divekar2024synthesizrrgeneratingdiversedatasets} can facilitate LLMs to generate more comprehensive answers that represent diverse perspectives. Yet, existing retrieval benchmarks do not focus on evaluating retrieval diversity. 

In this work, we study retrieval diversity, building a framework for evaluating whether a set of top-scored outputs from the retrieval system cover diverse \textit{perspectives}\footnote{We loosely define \textit{perspective} as a particular attitude or viewpoint expressed towards the input question.} for an input query.  ~\Cref{fig:intro} shows examples of possible perspectives for different input questions. To tackle this task, we curate a \textbf{Be}nchmark for \textbf{R}etrieval \textbf{D}iversity for \textbf{S}ubjective questions, \textbf{\bench}, which contains 3K complex questions, each paired with on average 2.3 perspectives. The questions come from three sources: survey question collection~\cite{santurkar2023whose}, debate question corpus~\cite{wachsmuth-etal-2018-retrieval}, and newly scrapped data from a debate platform (Kialo) with argument maps.






Existing retrieval benchmarks are either evaluated through string match to reference answers~\cite{Karpukhin2020DensePR} or exact match to the annotated document index in the corpus~\cite{Thakur2021BEIRAH,xiao2024rarbreasoningretrievalbenchmark}. We do not assume a corpus with gold document annotated and evaluate retrievers in an open-world setting (i.e., without assuming specific knowledge sources). To evaluate in this more realistic scenario, we develop an LLM-based automatic evaluator that determines if a document contains a certain perspective. We evaluate the performance of a retriever paired with a corpus. To quantify retrieval perspective diversity, we modify the evaluation metric used in multi-answer retrieval~\cite{min-etal-2021-joint}, where each question is paired with multiple valid factoid answers. \textsc{MRecall} @ $k$ measures whether a set of top-$k$ documents cover all valid perspectives. Figure~\ref{fig:intro} provides example instances in \textbf{\bench} and illustrates the proposed task. 




Having constructed an evaluation framework, we test popular retrievers~\cite{robertson2009probabilistic,karpukhin-etal-2020-dense,izacard2022unsupervised} and three corpora. Given the subjectivity of the questions we consider, Wikipedia alone is insufficient for answering many questions. Therefore, we use Sphere~\cite{piktus2021web}, a subset of a web snapshot (CCNet~\cite{wenzek-etal-2020-ccnet}), and a retrieval corpus built on the fly for each question from the Google search engine output. We find that pairing a dense retriever with web corpora (Sphere and Google Search) yields the most diverse outputs. Yet, our experimental result suggests that current retrievers surface relevant documents but cannot present document sets with diverse perspectives, even when retrieving from a richer web corpus. 



To enhance the diversity of the retrieval results, we implement simple re-ranking~\cite{carbonell1998use} and query expansion approaches~\cite{mao-etal-2021-generation}. The former re-computes the scores for each document by penalizing documents that are similar to the documents that were retrieved before. The latter first generates multiple perspectives regarding the question using LLM (GPT-4) and uses them to guide diverse retrieval. Encouragingly, both methods showed gains on the strongest dense base retriever~\cite{izacard2022unsupervised}, despite not boosting performances for all settings.  
We further provide rich analysis studying the coverage of each corpus, retriever sycophancy, and whether retrievers prefer supporting or opposing perspectives to the input query.


\ifacl@anonymize
 We will release all our code, data, and trained model checkpoints to promote future work in studying retrieval diversity for complex questions.
\else
All our code, data, and trained model checkpoints are released to promote future work in studying retrieval diversity for complex questions.\footnote{{\url{https://timchen0618.github.io/berds/}}}
\fi

\section{Related Work} \label{sec:related}

\paragraph{Diverse Perspectives}
Human society possesses pluralistic values~\cite{sorensen2024roadmap} and exhibits diverse perspectives in various scenarios, including responses to survey questions~\cite{santurkar2023whose,durmus2023towards}, news editorials~\cite{liu-etal-2021-multioped}, and stance with respect to debate or controversial questions~\cite{chen-etal-2019-seeing, chen-etal-2022-design}. \citet{Hayati2023HowFC} looks into eliciting diverse perspectives from LLM on social norm topics. \citet{wan2024evidence} studies LLM generation when in-context documents provide contradicting statements. Concurrent work~\cite{xu2024debateqa} studies evaluating perspective diversity of debatable questions. One way to encourage diversity in LLM output is retrieval augmentation, by providing documents with diverse perspectives. We focus on retrieving diverse opinions from existing corpus instaed of generating diverse opinions from LLMs.

\paragraph{Retrieval Diversity}

Previous works have investigated retrieval diversity in terms of entity ambiguity~\cite{clarke2008novelty, agrawal2009diversifying, wang2017search}. Closer to our definition of diversity, ~\citet{chen-etal-2019-seeing} propose a task \textit{substantiated perspective discovery}, where systems are supposed to discover diverse perspectives taking stances against a given claim. ~\citet{chen-etal-2022-design} further argues for the need of a multi-perspective search engine that provides direct and diverse responses. Despite these efforts, there is no benchmark for retrieving diverse perspectives in realistic corpora. We formalize such task and construct evaluation benchmark and metrics. Concurrent work \cite{ziems2024measuring} studies the bias of retrievers along a polar axis (e.g. capitalism vs. socialism) for debatable questions and develops an unsupervised automatic metric. 

Concurrent work~\cite{zhao2024beyond} explores the perspective-awareness of retrieval systems, where they repurpose existing datasets to build a retrieval benchmark where queries are supplemented with a perspective. While sharing a similar goal, they evaluate retrieval with perspectives as input and measure if systems follow them, and in our settings, the retrievers are not given target perspectives. Furthermore, our work expands knowledge sources significantly using a web corpus, while their work focuses on much smaller corpora.


\paragraph{Stance Detection}~\cite{mohammad-etal-2016-semeval,aldayel2021stance} evaluates if a piece of text supports a target (an entity, idea, opinion, claim, etc.). \citet{sen2018stance} investigate multi-perspective questions in health information and propose methods to classify search results according to their stance towards the question. \citet{wan2024evidence} determines the stance of search results to contentious questions using LLMs. Our perspective detection (\Cref{ssec:perspective-detection}) subsumes stance detection, which classifies the target as positive or negative (binary). Perspective detection can be generalized to any number or type of perspectives.


\vspace{-0.3em}


\paragraph{Retrieval from the Web}
~\citet{piktus2021web} proposes to retrieve from the web to account for the open-domain nature of real-world knowledge-intensive applications. Previous works explored retrieval with search engines (e.g. Google Search API) to help fact-checking~\cite{chen-etal-2024-complex}, open-domain question-answering~\cite{lazaridou2022internet}, answer-editing~\cite{gao-etal-2023-rarr}, and retrieval-augmented generation~\cite{li-etal-2023-web}. Other works have investigated attributing LM outputs to evidence documents retrieved from the web~\cite{malaviya2023expertqa,gao-etal-2023-enabling}. 


\section{Task Formulation}
\label{sec:task}

The retrieval system will be provided a question $q$ and a corpus $C$, and return a ranked list of $k$ documents, $D = \{d_1, d_2, ..., d_k\}$, where $d_i \in C$ and $d_i$ is the $i$-th most relevant document in the corpus to the question $q$.
\paragraph{Background: Multi-answer Retrieval} 

In earlier work~\cite{robertson2009probabilistic,voorhees1999trec,chen-etal-2017-reading,lee-etal-2019-latent}, each question $q$ is assumed to have a single gold answer $a$. Thus, the system is evaluated by assigning a binary label, \textsc{Recall} @ $k$, indicating if any document $d_i$ in the retrieved document set $D$ contains $a$. 

\citet{min-etal-2021-joint} extended retrieval task to cover questions with multiple valid answers. Here, each question $q$ is paired with gold answer set $\{a_1,... a_m\}$ where each answer $a_i$ is a short string. The goal is to retrieve a diverse set of documents such that the retrieval output will contain all $m$ answers. New evaluation metric \textsc{MRecall} @ $k$ is defined as (1) when number of answers $m > k$, \textsc{MRecall} @ $k$ = 1 if the retrieved document set $D$ contains $k$ answers. (2) when $m \leq k$, \textsc{MRecall} @ $k$ = 1 if $D$ contains all $m$ answers. Otherwise \textsc{MRecall} @ $k$ = 0. Simple string matching determines if a document $d_i$ contains answer $a$.

\begin{table}
\footnotesize
\begin{center}
\begin{tabular}{lccc}
\toprule
Dataset &  \# Total (Dev./Test) &  \# $p$ &  $|p|$ \\ \midrule

Arguana   & 1000 (250/750)& 2.0 & 12.71 \\

OpinionQA & 1176 (294/882)& 2.0 & 16.51 \\

Kialo     & 1032 (258/774) & 2.9 & 12.49 \\\midrule
All & 3208 (802 / 2407) & 2.28 & 13.84 \\

\bottomrule
\end{tabular}
\end{center}
\vspace{-1em}
\caption{Dataset statistics. The second column denotes the number of examples. We spare 25\% of each dataset as the development (Dev.) set, and use the remainder as the test set.  \# $p$ is the average number of perspectives per question, and $|p|$ denotes the average number of words (by NLTK \texttt{word\_tokenize}) in each perspective. }
\vspace{-0.7em}
\label{tab:data-stats}
\end{table}

\paragraph{Ours: Multi-perspective Retrieval} 
We consider questions on subjective topics that could be approached from different perspectives, without fixed or complete sets of reference answers. We assume a set of $m$ perspectives $\{p_1,..., p_m\}$ for addressing question $q$. Retrievers should output a diverse set of documents covering all $m$ perspectives.


\paragraph{Evaluation Challenges}Deciding if a document contains an answer can be approximated by simple string matching, but deciding if a document supports a perspective is nontrivial. The most accurate metric would be human judgments. However, this is not scalable, so we build a model approximating human judgment. We refer to this task of determining if a document contains a certain perspective as ``Perspective Detection'' (presented in \Cref{ssec:perspective-detection}).  We do not use off-the-shelf stance detection models~\cite{zhang2023investigating,alturayeif2023systematic} as the perspectives might not be binary. A single document could contain more than one perspective. 

\paragraph{Evaluation Metrics}
We use \textsc{MRecall} @ $k$ from prior work~\cite{min-etal-2021-joint}. The key difference is instead of checking whether a document contains any of answer $\{a_1,..., a_m\}$, we check whether a document supports any of perspectives $\{p_1, ..., p_m\}$. We also report \textsc{Precision} @ $k$, the percentage of documents in the retrieved document set $D$ that contain any perspectives. This measures the relevance of $D$ to the question.

{This evaluation setup does not assume annotated ``gold'' documents or answers for each question, enabling comparing different corpora. We discuss the effects of corpora in~\Cref{sec:analysis}.}







\section{The \bench{} Benchmark} \label{sec:benchmark}
The \bench{} Benchmark aims to measure retrieval diversity for questions that are opinionated or invite diverse perspectives. A single instance in \bench{} will consist of a question $q$ and a set of $m$ perspectives $\{p_1, p_2 ..., p_m\}$. Each perspective should be relevant to the question and distinct. We provide the data selection and construction process below. 

\subsection{Data Construction}

We aim to look for questions that elicit diverse valid perspectives naturally. 
Questions related to debate topics would be good candidates since they would induce at least two contradicting, valid viewpoints. Another natural choice is survey questions, where people would express differing opinions. 
We thus collect questions from a debate website (\url{kialo.com}) and a dataset of survey questions, OpinionQA~\cite{santurkar2023whose}. We also generate questions from Arguana~\cite{wachsmuth-etal-2018-retrieval}, a dataset of contradicting arguments. We provide examples for each dataset in \bench{} in~\Cref{fig:intro}, and data statistics in~\Cref{tab:data-stats}. For both Arguana and OpinionQA, we only consider two contradictory perspectives, one supporting and one opposing. These perspectives may not be exhaustive but cover a range of arguments for the question. \footnote{We mostly focus on binary perspectives since they are well-defined for contentious queries. Future work could explore more fine-grained and multi-dimensional perspectives.} We provide examples of the raw data and generated data in~\Cref{tab:dataset-ex-compare}.



\begin{table}
\footnotesize
\begin{center}
\begin{tabular}{lccc}
\toprule
Corpus & Source    & \# pas. &  \# doc.\\
\midrule
Wikipedia    & Wikipedia & 22 M    & 5.9 M  \\ 
Sphere  & CCNet     & 906 M   & 134 M  \\
\multicolumn{2}{c}{Google Search output}& 4461.7 &74.2\\
\bottomrule
\end{tabular}
\end{center}
\vspace{-0.8em}
\caption{Retrieval corpus statistics.  \# pas. is the number of passages; \# doc. is the number of documents, where each document is a 100-word passage. For Google Search output, we construct a corpus for each example, and we report the average statistics across examples.}
\vspace{-1.0em}
\label{tab:corpus-stats}
\end{table}

\paragraph{Arguana} is a dataset on the task of retrieving the best counterargument. As each instance in the Arguana dataset contains contradicting arguments (the query and its corresponding counterargument), it naturally induces two distinct perspectives. We generate questions from the contradicting arguments, and then generate corresponding perspectives for each argument, using GPT-4.

\paragraph{Kialo}
We collect 1,032 questions from a debate website (\url{kialo.com}), which mostly consist of Yes/No questions (91\%). For Yes/No questions, the website provides supporting statements converted from the questions. We consider the provided statement as one perspective and generate from it an opposing perspective using GPT-4. For the rest of the questions (9\%) that naturally invite more than two perspectives (e.g. What’s the best Twitter/X alternative?), we generate corresponding perspectives (e.g. The best Twitter/X alternative is Kialo) for each answer (e.g. Kialo) in the website. 

\paragraph{OpinionQA} is a dataset of survey questions on various topics including privacy, political views, and health, targeted toward the citizens of the United States. The dataset originates from the annual Pew American Trends Panel survey. 
The dataset contains questions asking about personal experiences (\textit{Regardless of whether or not you own a gun, have you ever fired a gun?}), which we filter by prompting GPT-4, removing 22.7\% of the questions (345 out of 1521 in total). The survey questions come with multiple options, with each representing a certain degree of support for the question. Conditioned on the survey question and the options, we convert each survey question into a natural language question and generate one supporting and one opposing perspective with GPT-4.

We use \texttt{gpt-4-0613} for all data generation. All the input prompts for generation can be found in~\Cref{ssec:prompt-data}. The total cost is about \$100 USD. {We manually examine 20 examples of each subset, and only two examples (out of 60) contain minor errors (see Appendix~\ref{ssec:manual-errors} for examples)}.


\vspace{-0.3em}
\subsection{Retrieval Corpus} 
\label{ssec:corpus}
We consider three types of retrieval corpus: Wikipedia, web snapshot, and the web itself. While prior work on factoid QA~\cite{chen-etal-2017-reading,lee-etal-2019-latent,karpukhin-etal-2020-dense} focused on retrieving from Wikipedia only, our task, designed to uncover a wide range of opinions and perspectives, will benefit from retrieving from a web corpus. 

\paragraph{Wikipedia}
We consider the Wikipedia dump processed by~\citet{karpukhin-etal-2020-dense}, consisting of disjoint 100-word segments. 

\paragraph{Web Snapshot: Sphere}
Prior works on retrieval usually adopt Wikipedia as the main retrieval corpus. However, Wikipedia alone has limited knowledge coverage~\cite{redi2020taxonomy}. This is even more likely in our case of complex and subjective queries, as Wikipedia is supposed to be neutral. One downside of moving from Wikipedia to a Web snapshot is potentially returning documents from less credible sources. We use the Sphere~\cite{piktus2021web} corpus, which is a subset of CCNet~\cite{wenzek-etal-2020-ccnet}, as the web knowledge source.

\paragraph{Online Web Search: Google Search API}
Instead of building a static corpus and retrieving documents from it, one can use a web search engine to retrieve new documents on the fly. This enables the system to access bigger, more up-to-date source documents. The downside of using web search API ~\cite{Nakano2021WebGPTBQ,Yoran2023MakingRL} is the lack of reproducibility.\footnote{Prior work~\cite{chen-etal-2024-complex} studied the reproducibility of search API results, comparing URLs between the sets of documents retrieved at different timestamps with the same query. They report only 30\% of URLs overlap when queried two months apart.}

We will use Google Search API to build web corpus on the fly.\footnote{\url{https://serper.dev/}} We first obtain the raw HTML of top hundred documents, which are then processed and converted to plain text using html2text\footnote{\url{https://pypi.org/project/html2text/}} and readability,\footnote{\url{https://github.com/buriy/python-readability}} following \citet{chen-etal-2022-generating, chen-etal-2024-complex}. We split the processed documents into 100-word segments and take all the segments as the corpus. Unlike Wikipedia and Sphere, we build a separate corpus for each question.

\begin{table*}
\footnotesize
\begin{center}
\vspace{-0.5em}
\begin{tabular}{ll cc cc cc |ccc}
\toprule
  \multirow{3}{*}{Corpus} &\multirow{3}{*}{Model}  & \multicolumn{6}{c|}{Datasets} & \multicolumn{3}{c}{Macro-Average}  \\
&&\multicolumn{2}{c}{Arguana} & \multicolumn{2}{c}{Kialo} & \multicolumn{2}{c|}{OpinionQA} &  \\
 &   & Prec. & MRec.  & Prec. & MRec.  & Prec. 
 & MRec.& Prec. & MRec. & Prec.@ 1 \\ \midrule 
 
\multirow{5}{*}{Wiki} & BM25&22.27&10.8&23.82&10.85&10.39&2.95&18.83&8.20&21.72\\
 &DPR&18.88&7.73&19.48&7.62&13.63&3.29&17.33&6.21&20.22\\
 & \textsc{Contriever}&35.79&19.2&32.97&15.63&23.99&8.28&30.92&14.37&36.85\\ 
 & \textsc{Tart} & 37.84 & 21.07 & 33.97 & 16.80 & 25.36 & 8.40 & 32.39	& 15.42 & 38.62\\ 
 & NV-Embed-v2 & 40.69 & 23.20 & 38.99 & 22.27 & 28.80 & 8.13 & 36.16 & 17.87 & 43.15 \\

\midrule
 \multirow{4}{*}{Sphere} & BM25&44.96&27.6&43.33&24.16&50.11&22.45&46.13&24.74&51.26 \\
 & DPR&10.80&4.00&13.85&5.43&8.34&2.15&11.00&3.86&11.65\\
 & 
\textsc{Contriever}&58.21&38.27&54.96&28.94&52.77&24.72&55.31&30.64&58.67 \\ 
& \textsc{Tart} &  65.47 &  43.60 & 56.47 & 32.80 &  63.25 &  31.60 &  61.73	&  36.00 & 65.16\\
& NV-Embed-v2  &  \textbf{69.84} &  \textbf{43.87} & \textbf{66.85} & \textbf{40.31} &  \textbf{64.94} &  \textbf{34.35} &  \textbf{67.21}	& \textbf{39.51} & \textbf{71.27}\\

\midrule
 & BM25&35.15&21.07&35.43&22.61&41.27&21.09&37.28&21.59&43.02\\
  Google Search &\textsc{Contriever}&52.32&27.6&50.31&28.94&48.91&22.79&50.51&26.44 &57.56 \\
 Output& \textsc{Tart} & 53.33 & 30.13 & 50.48 & 27.87 & 50.83 & 24.67 & 51.55 & 27.56 & 59.07\\
 & NV-Embed-v2 & 62.40 & 39.87 &  64.19 &  37.60 & 56.88 & 28.67 & 61.16 & 35.38  &  67.87\\

\bottomrule
\end{tabular}
\end{center}\vspace{-1em}
\caption{Performance of retrievers on the test split of \bench. We do not report NV-Embed-v2 results on Sphere for computational limitations. Overall, retrievers show better performance with Sphere corpus, and NV-Embed-v2 shows the strongest performance in both corpora that it was evaluated on.}\vspace{-0.5em}
\label{tab:main-results}
\end{table*}

\section{Perspective Detection}
\label{ssec:perspective-detection}

We aim to build evalution metrics for retrieval diversity. 
Unlike factoid QA, where we can evaluate whether a document contains an answer by string match~\cite{lee-etal-2019-latent, min2021neurips, min-etal-2021-joint}, evaluating whether a document contains a perspective is nontrivial, a research question on its own~\cite{sen2018stance, aldayel2021stance}. We need a system that identifies whether the retrieved document set contains a perspective. We define the subtask as ``Perspective Detection'', and detail the process of building an efficient model for this subtask.

\paragraph{Task and Metric}

Given a perspective and a document, the model assigns a binary label $y$ indicating whether the document contains the perspective. Then, we measure the performance of a model by comparing its prediction with the reference label, measuring accuracy and F1 score. 

\paragraph{Reference Annotation}
\label{ssec:construct-per}

For each document-perspective pair, we consider human judgment as the gold label. 
The human-labeled test set contains 542 document-perspective pairs, which are annotated by the authors of this paper. We compute inter-annotator agreement on 176 examples, yielding a Cohen Kappa agreement of 0.56. 16.8\% of the data receive a positive label. The examples are taken from top five retrieval results from the Sphere corpus with BM25~\cite{robertson2009probabilistic} and DPR~\cite{karpukhin-etal-2020-dense} retriever described in~\Cref{ssec:baselines}, and equally split between three datasets. Details on the construction of the human reference set can be found in~\Cref{ssec:pers-det-test}.

\subsection{Evaluated Approaches}
\paragraph{Random} We report a baseline that randomly assign predictions $y= 1$ or $0$ according to the label distribution of each dataset. 

\paragraph{Off-the-shelf NLI models} 
We could interpret ``a document $d_i$ contains perspective $p_j$" as ``a premise $d_i$ supports a hypothesis $p_j$". Then it becomes natural to use NLI models to identify whether a perspective is represented in a document. We use a T5-11B model provided by~\citet{honovich-etal-2022-true-evaluating}. 

\paragraph{Prompted LMs}
Instead of using an NLI model, we can prompt LLMs to decide whether a document contains a perspective. From the open-source LMs, we choose Llama-2-chat~\cite{touvron2023llama}, an instruct-tuned version of Mistral-7B~\cite{jiang2023mistral}, Zephyr-7B~\cite{tunstall2023zephyr}, and Gemma-7B~\cite{team2024gemma}. Lastly, we use GPT-4 (\texttt{gpt-4-0613}). For GPT-4, the best-performing model, we further experiment with prompting with one in-context example. 




\paragraph{Fine-tuned LMs} Prompting GPT-4 shows promising results in this task but requires additional API calls, which incur extra costs whenever we want to evaluate new systems on \bench{}. We thus fine-tune the Mistral-7B model with GPT-4 predictions with one in-context example as labels. The fine-tuning dataset contains 3K examples corresponding to 50 unique questions, and implementation details can be found in~\Cref{ssec:finetune}. 


\subsection{Results}

\begin{table}
\footnotesize
\begin{center}
\begin{tabular}{lrrr}
\toprule

Models & \% Pos & Acc & F1 \\ \midrule
Reference human  & 16.8 & - & - \\
Another human & 22.7 & 85.2 & 65.8 \\
Random & 16.8 & 72.1 & 16.9 \\ \midrule
NLI model (T5-11B) & 7.2 & 85.2 & 38.5 \\
Llama-2-13b-chat (zero-shot) & 31.1 & 70.7 & 42.6 \\
Llama-2-70b-chat (zero-shot) & 27.4&77.7& 52.6\\ 

Zephyr (zero-shot) & 22.4 & 80.6	&53.3 \\ 
Gemma (zero-shot)  & 58.4 & 48.9 & 37.2 \\ 

Mistral-7B (zero-shot) & 19.2 &82.2 & 52.0 \\

GPT-4 (zero-shot) & 17.4 & \textbf{87.6}	& \textbf{65.6} \\ \midrule

GPT-4 (one-shot) & 15.4 &  \textbf{89.1}	& \textbf{67.8}\\
Mistral-7B (fine-tuned) & 15.8  & 87.6 & 62.2 \\

\bottomrule
\end{tabular}
\end{center} \vspace{-0.5em}
\caption{Perspective detection results on human-labeled data (n=542). ``Another human'' performance is computed by taking one set of human labels as the ground truth and the other as predictions on the 176 examples where we compute inter-annotator agreement. }
\vspace{-1em}
\label{tab:pers-det}
\end{table}
We report the results on the human-labeled document-perspectives pairs in~\Cref{tab:pers-det}. The off-the-shelf entailment model (T5) does not work well, whereas GPT-4 is the best model. Fine-tuned Mistral-7B obtains the best results in open-source models, surpassing Llama-2-70b-chat. The prediction from fine-tuned Mistral-7B matches that of GPT-4 on 94\% of instances.


As the fine-tuned Mistral-7B comes close to GPT-4 performance, we use this model as the automatic metric for evaluating retrieval diversity, providing reproducible, cheaper evaluation.

%





\section{Retrieval Approaches} \label{sec:exp}
We consider a suite of retrievers (sparse and dense) on our dataset. As these retrievers do not encourage the diversity of their output, we implement a re-ranking method and a query expansion method.





\paragraph{Base Models} \label{ssec:baselines}

We consider one sparse retriever (BM25) and four dense retrievers, and we describe them below.
\begin{itemize}[noitemsep,leftmargin=10px]
    \item BM25~\cite{robertson2009probabilistic}: a widely adopted bag-of-word retrieval function considering token-matching between questions and documents. We use the Pyserini package for the Wikipedia corpus. For the Sphere corpus, we use the index provided by ~\citet{piktus2021web}. 
    \item DPR~\cite{karpukhin-etal-2020-dense}: a dual encoder trained with contrastive learning on the Natural Questions (NQ) dataset~\cite{kwiatkowski-etal-2019-natural}.
    \item \textsc{Contriever}~\cite{izacard2022unsupervised}: an encoder trained with a self-supervised contrastive objective.\footnote{DPR uses separate encoders for document and question embeddings, and \textsc{Contriever} uses a shared encoder. Both models are initialized from a BERT-base encoder.} We use the version that is further fine-tuned on the MSMARCO dataset~\cite{nguyen2016ms} after self-supervised learning, which performed best in the original paper.
    \item \textsc{Tart}~\cite{asai-etal-2023-task}: an encoder trained to make use of instruction in addition to the query. We implement \textsc{Tart}-full, the cross-encoder version, and the top hundred documents from \textsc{Contriever} are used as initial candidates. 
    \item NV-Embed-v2~\cite{lee2024nv}: the state-of-the-art encoder.\footnote{Ranked first in the MTEB~\cite{muennighoff2022mteb} benchmark as of Oct 15, 2024.} We do not include results on Sphere due to constraints on the compute. Other BERT-based encoder models (110M parameters) uses an embedding size of 768, while NV-Embed-v2 model (7B parameters) uses an embedding size of 4096, significantly increasing the cost of building document indices.
\end{itemize}


\paragraph{Re-ranking}
\label{ssec:reranking}
We apply the Maximal Marginal Relevance (MMR) method~\cite{carbonell1998use}, which uses the following selection criteria for re-ranking:
\begin{equation} \label{eq:mmr}
\footnotesize
     \argmax_{D_{i} \in R \setminus S}[ \lambda Sim_1(D_i, Q) - (1-\lambda) \max_{D_j \in S} Sim_2(D_i, D_j) ]
\end{equation}
\noindent where $R$ denotes a set of top retrieved documents considered for re-ranking ($|R|=100$ in our case), $S$ denotes the document set that is already selected, and $\lambda$ is a hyperparameter, which we tune on the development split.\footnote{We normalize the $Sim_1$ scores to be between 0 and 1, and consider $\lambda$ values of [0.5, 0.75, 0.9, 0.95, 0.99].} $Sim_1$ is the score from the retriever. $Sim_2$ is the cosine similarity between two document embeddings computed using the Universal AnglE Embedding~\cite{li2023angle}. We choose this model as it is the top-ranked open model on the MTEB leaderboard~\cite{muennighoff-etal-2023-mteb} with less than 1B parameters (335M). {Compact models are preferable as we compute document embeddings for the top hundred documents per example.} The experimental details can be found in~\Cref{ssec:rerank-details}.

\paragraph{Query Expansion} \label{ssec:div-aware}

Prior work on factoid QA~\cite{mao-etal-2021-generation} showed that first generating an answer with LLM and then using the generated answer to augment the question can improve retrieval. Similarly, we first generate multiple ($n$) perspectives on the given question using LLMs (\texttt{gpt-4-0613}), and query the retriever with each generated perspective.\footnote{$n$ is not predetermined; GPT-4 could generate varying numbers of perspectives. $n$ = 6.1 on average.} Ground truth perspectives are not provided in the prompt. We take the top document from each document set retrieved using the generated perspectives in a round-robin fashion $k$ times to form a list of $n*k$ documents, following the order in which the perspectives are generated. We then take the top $k$ unique documents from this list, skipping duplicates, as the same documents could be obtained from different queries. The details of query expansion step can be found in ~\Cref{ssec:prompt-div} and ~\Cref{ssec:ex-diversity-aware}.  

\begin{table*}
\footnotesize
\begin{center}
\vspace{-0.5em}
\begin{tabular}{ll cc cc cc}
\toprule
\multirow{2}{*}{Corpus} &\multirow{2}{*}{Retriever}  &\multicolumn{2}{c}{Default} & \multicolumn{2}{c}{+ Re-ranking} & \multicolumn{2}{c}{+ Query Expansion}    \\
&   & Prec. & MRec.  & Prec. & MRec.  & Prec. 
 & MRec.\\ \midrule 
 
\multirow{3}{*}{Wiki} & BM25&18.83&8.20& 18.66 (-0.9\%)&8.03 (-2.1\%) &\textbf{21.31} (\color{blue}{+13.2\%})&\textbf{10.39} (\color{blue}{+26.7\%}) \\
 & DPR&17.33&6.21&16.62 (-4.1\%)&6.06 (-2.5\%)&\textbf{17.78} (\color{blue}{+2.6\%})&\textbf{6.65} (\color{blue}{+7.1\%}) \\
 & \textsc{Contriever}&30.92&14.37&27.89 (-9.8\%)&\textbf{15.54} (\color{blue}{+8.2\%})&\textbf{33.23} (\color{blue}{+7.5\%})&14.84 (\color{blue}{+3.3\%})\\

\midrule
 \multirow{3}{*}{Sphere} & BM25&\textbf{46.13}&24.74&45.43 (-1.5\%)&25.38 (\color{blue}{+2.6\%})&44.16 (-4.3\%)&\textbf{26.24} (\color{blue}{+6.1\%})\\
 & DPR&\textbf{11.00}&\textbf{3.86}&8.16 (-25.8\%)&2.60 (-32.6\%)&6.20 (-43.6\%)&2.01 (-48.0\%)\\
 & \textsc{Contriever}&\textbf{55.31}&30.64&54.39 (-1.7\%)&33.55 (\color{blue}{+9.5\%})&54.91 (-0.7\%)&\textbf{33.74} (\color{blue}{+10.1\%})\\ 

\midrule
 Google Search & BM25&37.28&21.59&36.75 (-1.4\%)&21.18 (-1.9\%)&\textbf{44.87} (\color{blue}{+20.3\%})&\textbf{24.96} (\color{blue}{+15.6\%}) \\ 
 Output& \textsc{Contriever}&50.51&26.44&49.33 (-2.3\%)&\textbf{26.56} (\color{blue}{+0.4\%})&\textbf{52.60} (\color{blue}{+4.1\%})&24.83 (-6.1\%) \\ 

\bottomrule
\end{tabular}
\end{center} \vspace{-0.5em}
\caption{Performances on \bench{} after re-ranking and query expansion. Each cell reports the macro-average over three datasets in \bench. \textbf{Bolded} numbers are the best performance in each row. {\color{blue}{Blue}} numbers indicate increases compared to the default setting. }
\label{tab:improve-div}
\vspace{-1em}
\end{table*}

\section{Experimental Results} \label{sec:results}
\vspace{-0.1em}
We report \textsc{MRecall} @ $5$ and \textsc{Precision} @ $5$,\footnote{We additionally report the results for $k$ = 10 in~\Cref{ssec:results-k10}, where very similar trends hold.} as defined in~\Cref{sec:task}. 
All evaluation metrics require judging whether each retrieved document contains a perspective, or perspective detection (\Cref{ssec:perspective-detection}). We use a fine-tuned Mistral-7B model as the evaluator.\footnote{We will open-source this model to promote future work.}


\Cref{tab:main-results} presents the performance of base retrievers on \bench{} using three retrieval corpus described in~\Cref{ssec:corpus}. We do not report DPR results on Google Search outputs as we find it does not generalize well beyond the Wikipedia corpus. We do not report NV-Embed-v2 results on Sphere due to limited compute. 

Unlike prior work~\cite{karpukhin-etal-2020-dense} which assumes a gold corpus, our retrieval corpus might not contain documents representing diverse perspectives on the provided question. Thus, the performance metrics reflect both the coverage of the corpus and the retriever's ability to surface documents supporting diverse perspectives. We analyze the coverage of each corpus in~\Cref{sec:analysis}.


\paragraph{Effect of Corpus.}
Comparing retrievers operating on Wikipedia and Sphere, most retrievers achieve higher \textsc{MRecall} and \textsc{Precision} when retrieving from Sphere, except DPR. We hypothesize that DPR is trained only on the documents from the Wikipedia corpus so that it is not generalizing to the web corpus. 
Taking web search results (Google Search) as a corpus shows improved diversity over Wikipedia, but lags behind retrieving from the entire web (Sphere). The gap in the corpus size might cause this -- we only take the top 100 Google Search results and build corpora with them (resulting in a corpus size of 4.4K passages on average), while Sphere contains 906M passages.

\paragraph{Comparing Retrievers.}
Across all experimental settings, \textsc{Tart} achieves the highest diversity, followed by \textsc{Contriever}, BM25, and DPR. BM25 generalizes better than DPR, which is fine-tuned on NQ with Wikipedia corpus. \textsc{Tart} achieves higher diversity than \textsc{Contriever}, suggesting that specifying the task with instructions helps diversity. Even the best setting (\textsc{Tart} on Sphere) only achieves an average \textsc{MRecall} of 32.8. This indicates even the best retriever struggles to retrieve a comprehensive set of documents. {We investigate the number of documents needed to be retrieved to cover diverse perspectives in~\Cref{ssec:rank}.} \textsc{Precision} @ 1 is as high as 58.67, showing that retrievers could surface one of the perspectives. 




\paragraph{Results: Approaches to Improve Diversity}
We report the results of re-ranking and query expansion approaches in~\Cref{tab:improve-div}. We report the results with three retrievers (BM25, DPR, Contriever), excluding TART and NV-Embed2 for computational costs. Re-ranking improves diversity on four of eight retriever-corpus settings, at the cost of \textsc{Precision} for all settings. The limited performance gains could be because semantic dissimilarity does not always infer differing perspectives. Query expansion improves the \textsc{Precision} for five settings and boosts \textsc{MRecall} for six settings (over 10\% for three of them). {This indicates that GPT-4 could diversify the search space by providing queries hinting different perspectives to the retrievers.} Future work can explore combining both approaches. 

\section{Analysis}\label{sec:analysis}

\paragraph{Do retrievers prefer documents with supporting or opposing perspectives to the question?}

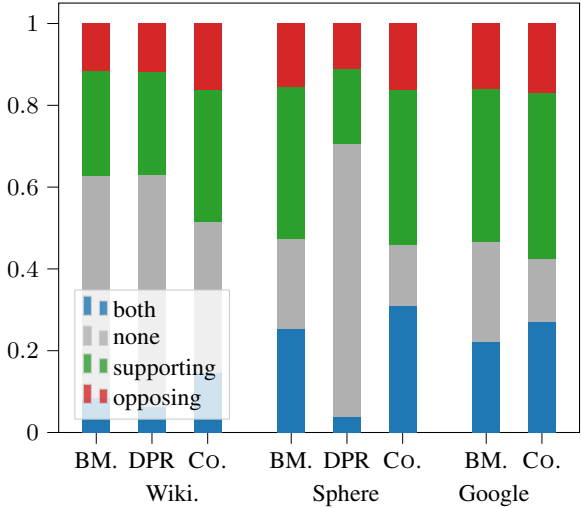
\begin{figure}[t!] 
  \centering \small
  
  \resizebox{0.95\columnwidth}{!}{
  \begin{tikzpicture}
  
  \definecolor{crimson2143940}{RGB}{214,39,40}
  \definecolor{darkgray176}{RGB}{176,176,176}
  \definecolor{darkorange25512714}{RGB}{176,176,176}
  \definecolor{forestgreen4416044}{RGB}{44,160,44}
  \definecolor{lightgray204}{RGB}{204,204,204}
  \definecolor{steelblue31119180}{RGB}{31,119,180}

\begin{axis}[
legend cell align={left},
legend style={
  fill opacity=0.8,
  draw opacity=1,
  text opacity=1,
  at={(0.03,0.03)},
  anchor=south west,
  draw=lightgray204
},
tick align=outside,
tick pos=left,
x grid style={darkgray176},
xlabel={\quad  \quad \textbf{Wiki}  \quad  \quad \quad \quad \textbf{Sphere}     \quad \quad \quad \textbf{Google API}},
xmin=-1.015, xmax=20.215,
xtick style={color=black},
xtick={0.3,2,3.7,5.4,7.9,9.6,11.3,13,15.5,17.2,18.9},
xticklabels={
  BM.,
  DP.,
  \textsc{Co.},
  \textsc{Ta.},
  BM.,
  DP.,
  \textsc{Co.},
  \textsc{Ta.},
  BM.,
  \textsc{Co.},
  \textsc{Ta.}
},
y grid style={darkgray176},
ymin=0, ymax=1.05,
ytick style={color=black}
]
\draw[draw=none,fill=steelblue31119180] (axis cs:-0.05,0) rectangle (axis cs:0.65,0.0845419222772521);
\addlegendimage{ybar,ybar legend,draw=none,fill=steelblue31119180}
\addlegendentry{both}

\draw[draw=none,fill=steelblue31119180] (axis cs:1.65,0) rectangle (axis cs:2.35,0.063649898612262);
\draw[draw=none,fill=steelblue31119180] (axis cs:3.35,0) rectangle (axis cs:4.05,0.146357462972599);
\draw[draw=none,fill=steelblue31119180] (axis cs:5.05,0) rectangle (axis cs:5.75,0.160889839599674);
\draw[draw=none,fill=steelblue31119180] (axis cs:7.55,0) rectangle (axis cs:8.25,0.241049421972125);
\draw[draw=none,fill=steelblue31119180] (axis cs:9.25,0) rectangle (axis cs:9.95,0.0318110055479542);
\draw[draw=none,fill=steelblue31119180] (axis cs:10.95,0) rectangle (axis cs:11.65,0.309094208158555);
\draw[draw=none,fill=steelblue31119180] (axis cs:12.65,0) rectangle (axis cs:13.35,0.347413291536319);
\draw[draw=none,fill=steelblue31119180] (axis cs:15.15,0) rectangle (axis cs:15.85,0.221725694243703);
\draw[draw=none,fill=steelblue31119180] (axis cs:16.85,0) rectangle (axis cs:17.55,0.27089100871699);
\draw[draw=none,fill=steelblue31119180] (axis cs:18.55,0) rectangle (axis cs:19.25,0.283587956806896);
\draw[draw=none,fill=darkorange25512714] (axis cs:-0.05,0.0845419222772521) rectangle (axis cs:0.65,0.629036189105797);
\addlegendimage{ybar,ybar legend,draw=none,fill=darkorange25512714}
\addlegendentry{none}

\draw[draw=none,fill=darkorange25512714] (axis cs:1.65,0.063649898612262) rectangle (axis cs:2.35,0.631784111567195);
\draw[draw=none,fill=darkorange25512714] (axis cs:3.35,0.146357462972599) rectangle (axis cs:4.05,0.515266582301791);
\draw[draw=none,fill=darkorange25512714] (axis cs:5.05,0.160889839599674) rectangle (axis cs:5.75,0.513997623223038);
\draw[draw=none,fill=darkorange25512714] (axis cs:7.55,0.241049421972125) rectangle (axis cs:8.25,0.461429894715204);
\draw[draw=none,fill=darkorange25512714] (axis cs:9.25,0.0318110055479542) rectangle (axis cs:9.95,0.698559450472036);
\draw[draw=none,fill=darkorange25512714] (axis cs:10.95,0.309094208158555) rectangle (axis cs:11.65,0.458967245443975);
\draw[draw=none,fill=darkorange25512714] (axis cs:12.65,0.347413291536319) rectangle (axis cs:13.35,0.464555523436948);
\draw[draw=none,fill=darkorange25512714] (axis cs:15.15,0.221725694243703) rectangle (axis cs:15.85,0.46670313627497);
\draw[draw=none,fill=darkorange25512714] (axis cs:16.85,0.27089100871699) rectangle (axis cs:17.55,0.42465576775249);
\draw[draw=none,fill=darkorange25512714] (axis cs:18.55,0.283587956806896) rectangle (axis cs:19.25,0.431573867444365);
\draw[draw=none,fill=forestgreen4416044] (axis cs:-0.05,0.629036189105797) rectangle (axis cs:0.65,0.883971966879292);
\addlegendimage{ybar,ybar legend,draw=none,fill=forestgreen4416044}
\addlegendentry{first}

\draw[draw=none,fill=forestgreen4416044] (axis cs:1.65,0.631784111567195) rectangle (axis cs:2.35,0.8814173556108);
\draw[draw=none,fill=forestgreen4416044] (axis cs:3.35,0.515266582301791) rectangle (axis cs:4.05,0.836304338795638);
\draw[draw=none,fill=forestgreen4416044] (axis cs:5.05,0.513997623223038) rectangle (axis cs:5.75,0.862404270490066);
\draw[draw=none,fill=forestgreen4416044] (axis cs:7.55,0.461429894715204) rectangle (axis cs:8.25,0.840249428824095);
\draw[draw=none,fill=forestgreen4416044] (axis cs:9.25,0.698559450472036) rectangle (axis cs:9.95,0.881370933516623);
\draw[draw=none,fill=forestgreen4416044] (axis cs:10.95,0.458967245443975) rectangle (axis cs:11.65,0.836787017230562);
\draw[draw=none,fill=forestgreen4416044] (axis cs:12.65,0.464555523436948) rectangle (axis cs:13.35,0.842506955819785);
\draw[draw=none,fill=forestgreen4416044] (axis cs:15.15,0.46670313627497) rectangle (axis cs:15.85,0.839074384553947);
\draw[draw=none,fill=forestgreen4416044] (axis cs:16.85,0.42465576775249) rectangle (axis cs:17.55,0.830678803217856);
\draw[draw=none,fill=forestgreen4416044] (axis cs:18.55,0.431573867444365) rectangle (axis cs:19.25,0.843269348784525);
\draw[draw=none,fill=crimson2143940] (axis cs:-0.05,0.883971966879292) rectangle (axis cs:0.65,1);
\addlegendimage{ybar,ybar legend,draw=none,fill=crimson2143940}
\addlegendentry{second}

\draw[draw=none,fill=crimson2143940] (axis cs:1.65,0.8814173556108) rectangle (axis cs:2.35,1);
\draw[draw=none,fill=crimson2143940] (axis cs:3.35,0.836304338795638) rectangle (axis cs:4.05,1);
\draw[draw=none,fill=crimson2143940] (axis cs:5.05,0.862404270490066) rectangle (axis cs:5.75,1);
\draw[draw=none,fill=crimson2143940] (axis cs:7.55,0.840249428824095) rectangle (axis cs:8.25,1);
\draw[draw=none,fill=crimson2143940] (axis cs:9.25,0.881370933516623) rectangle (axis cs:9.95,1);
\draw[draw=none,fill=crimson2143940] (axis cs:10.95,0.836787017230562) rectangle (axis cs:11.65,1);
\draw[draw=none,fill=crimson2143940] (axis cs:12.65,0.842506955819785) rectangle (axis cs:13.35,1);
\draw[draw=none,fill=crimson2143940] (axis cs:15.15,0.839074384553947) rectangle (axis cs:15.85,1);
\draw[draw=none,fill=crimson2143940] (axis cs:16.85,0.830678803217856) rectangle (axis cs:17.55,1);
\draw[draw=none,fill=crimson2143940] (axis cs:18.55,0.843269348784525) rectangle (axis cs:19.25,1);
\end{axis}

\end{tikzpicture}}
\vspace{-0.3em}
\caption{We report whether retrievers prefer the supporting or the opposing perspectives. Results are computed on the top-5 documents aggregated over three datasets. All retrievers favor the supporting perspectives. BM. denotes BM25 retriever, DP. denotes DPR, and \textsc{Co.} denotes \textsc{Contriever}. \textsc{Ta.} denotes \textsc{Tart} and Wiki. denotes the Wikipedia corpus.}\vspace{-0.5em}
    \label{fig:leaning}

    \end{figure}

Retrievers often fail to surface documents that cover comprehensive perspectives. Can we characterize the perspective that is more likely to be surfaced? We focus on questions with only two perspectives, one taking the supporting stance towards the question and the other opposing.\footnote{We identify this by prompting GPT-4; details and prompts can be found in~\Cref{ssec:details-sup-op}.}
We report whether the top five documents obtained by the retrievers cover only supporting, opposing, both, or neither perspectives in~\Cref{fig:leaning}. We observe that all retrievers retrieve the supporting perspectives significantly more often when they fail to retrieve both.

\paragraph{Retriever Sycophancy.}
{Sycophancy refers to one's tendency to tailor their response to please their interaction partner. Prior work has explored how LLMs exhibit sycophancy~\cite{perez-etal-2023-discovering}. We examine if similar behavior is observed with the retrievers; i.e. do retrievers favor documents that share the perspective with the question? }
For example, would the retrievers surface more documents supporting a perspective, ``ChatGPT will do more harm than good.'', if the perspective is used as the question for the retrievers, compared to using ``Chat GPT will do more good than harm" as the question? 
We conduct a controlled study where we input either the supporting or opposing perspectives as the query to the retriever, and observe the changes to their leaning.

\begin{figure}[tb!] 
\centering \small
\resizebox{0.95\columnwidth}{!}{
\begin{tikzpicture}

\definecolor{darkgray176}{RGB}{176,176,176}
\definecolor{darkorange25512714}{RGB}{230, 173, 140}
\definecolor{forestgreen4416044}{RGB}{230, 199, 140}
\definecolor{lightgray204}{RGB}{204,204,204}
\definecolor{steelblue31119180}{RGB}{189, 199, 133}
\begin{axis}[
legend cell align={left},
legend style={
  fill opacity=0.8,
  draw opacity=1,
  text opacity=1,
  at={(0.03,0.03)},
  anchor=south west,
  draw=lightgray204
},
tick align=outside,
tick pos=left,
x grid style={darkgray176},
xmin=0, xmax=70,
xtick style={color=black},
y grid style={darkgray176},
ymin=-0.3125, ymax=3.8125,
ytick style={color=black},
ytick={0.25,1.25,2.25,3.25},
yticklabels={Wiki\_BM25,Wiki\_DPR,Wiki\_\textsc{Cont.},Wiki\_\textsc{Tart}}
]
\draw[draw=none,fill=steelblue31119180] (axis cs:0,-0.125) rectangle (axis cs:43.1297551007088,0.125);
\addlegendimage{ybar,ybar legend,draw=none,fill=steelblue31119180}
\addlegendentry{supporting perspective}

\draw[draw=none,fill=steelblue31119180] (axis cs:0,0.875) rectangle (axis cs:46.7326978392602,1.125);
\draw[draw=none,fill=steelblue31119180] (axis cs:0,1.875) rectangle (axis cs:38.0968202889623,2.125);
\draw[draw=none,fill=steelblue31119180] (axis cs:0,2.875) rectangle (axis cs:43.5836965420206,3.125);
\draw[draw=none,fill=darkorange25512714] (axis cs:0,0.125) rectangle (axis cs:41.2318349221544,0.375);
\addlegendimage{ybar,ybar legend,draw=none,fill=darkorange25512714}
\addlegendentry{default query}

\draw[draw=none,fill=darkorange25512714] (axis cs:0,1.125) rectangle (axis cs:42.6292828884819,1.375);
\draw[draw=none,fill=darkorange25512714] (axis cs:0,2.125) rectangle (axis cs:34.1715347256307,2.375);
\draw[draw=none,fill=darkorange25512714] (axis cs:0,3.125) rectangle (axis cs:41.8857448806255,3.375);
\draw[draw=none,fill=forestgreen4416044] (axis cs:0,0.375) rectangle (axis cs:35.3342111051821,0.625);
\addlegendimage{ybar,ybar legend,draw=none,fill=forestgreen4416044}
\addlegendentry{opposing perspective}

\draw[draw=none,fill=forestgreen4416044] (axis cs:0,1.375) rectangle (axis cs:21.9669721267659,1.625);
\draw[draw=none,fill=forestgreen4416044] (axis cs:0,2.375) rectangle (axis cs:20.1879064264385,2.625);
\draw[draw=none,fill=forestgreen4416044] (axis cs:0,3.375) rectangle (axis cs:26.4257839020671,3.625);
\end{axis}

\end{tikzpicture}}\vspace{-0.8em}
\caption{Retriever sycophancy. X-axis reports the propensity of retrieval output to prefer positive, the $\Delta=\frac{p-n}{p}$, where $p$ is the number of retrieved documents that contain supporting perspective and $n$ is the number of retrieved documents that contain opposing perspective. We label each setting as \texttt{[corpus]\_[retriever]}. In all retrievers, using supporting perspective as query leads to higher supportive leaning, and using opposing perspective as query leads to lower supportive leaning.  }
\label{fig:sycophancy}
\end{figure}
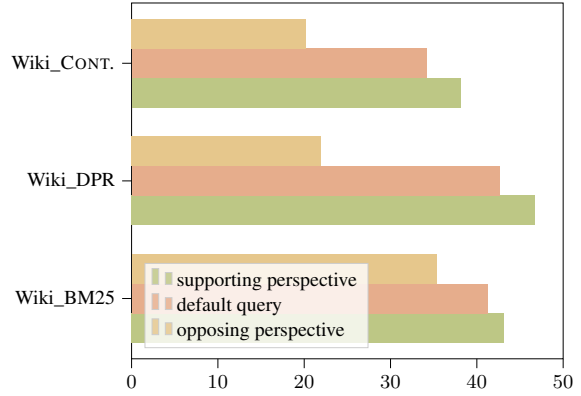

We quantify leaning towards supporting perspectives to be $\Delta$ = ($p$ - $n$)/$p$, where $p$ and $n$ are the percentages of documents containing supporting and opposing perspectives, respectively. Greater $\Delta$ indicates the retrievers favor the supporting perspectives more.  We present the results on the top 5 documents retrieved from the Wikipedia corpus in~\Cref{fig:sycophancy}.\footnote{Results on the other corpora show the same trends, except for DPR on the web corpus.} Querying the retrievers with supporting perspectives increases $\Delta$ across the board, and retrieving with opposing perspectives decreases $\Delta$. Retrievers tend to favor perspectives that they are prompted with. This partially explains the gains from the query expansion.

\paragraph{Is retrieval diversity limited by the corpora?} Unlike prior work, we do not assume a gold corpus that contains target information. The best \textsc{MRecall} @ $5$ on \bench{} achieved by any approach on average is below 40. 
How much of the failure is caused by the limitation of the corpus or that of retrievers? We answer this question by computing \textsc{MRecall} over a larger document set, where we combine the top 100 retrieved documents from all retrievers except NV-Embed-v2 for each corpus. If this number nears 100, there exists an optimal subset of $k$ documents for each question that achieve \textsc{MRecall} @ $k$ close to 1. We present the results in~\Cref{tab:upperbound}. Only the Wikipedia corpus contains insufficiently diverse information, achieving \textsc{MRecall} of 62\% when combining retriever outputs. In contrast, the web corpora do not limit retrievers' performances in diversity. This result hints at great room for improvement for the retrievers.

\begin{table}
\footnotesize
\begin{center}
\begin{tabular}{l cccc}
\toprule
Corpus & Arguana & Kialo & OpinionQA & Average \\
\midrule

{Wiki.} & 71.87 &61.24&55.10&62.74\\
{Sphere} & 92.80 &85.27&89.46&89.18\\
{Google} & 82.67 &82.69&77.66&81.01\\
\bottomrule
\end{tabular}
\end{center}\vspace{-1.0em}
\caption{\textsc{MRecall} over the union of the top 100 retrieved documents from all retrievers except NV-Embed-v2 for each corpus. These values are substantially higher than \textsc{MRecall} @ $5$, suggesting the corpora (especially Sphere) have good information coverage.   }
\label{tab:upperbound}
\end{table}


\section{Conclusion}
We study retrieval diversity for complex, subjective questions. 
We curate a benchmark \bench{} that evaluates retrieval diversity, provides automatic metrics, and establishes the performances of retriever baselines. {Our experiments show that current retrievers cannot retrieve comprehensive information in the proposed setting.}

After retrieving diverse documents, summarizing such rich information into a coherent response~\cite{laban-etal-2022-discord,huang2023embrace,zhang2023fair} could be useful for the users. 

\section*{Limitations}
\begin{itemize}[noitemsep,leftmargin=10px]
    \item The perspective set in our dataset is mostly binary (the average number of perspectives is 2.9 for Kialo and 2.0 for Arguana and OpinionQA). More fine-grained perspectives could be explored (e.g. reasons for supporting positive stance).  
    \item Our dataset covers debate and opinion survey questions. Exploring other domains like healthcare would be an interesting future direction.
    \item The evaluator LM has 7B parameters, making evaluation slow. Evaluating one dataset in \bench{} using the evaluator takes about three hours on an NVIDIA A40 GPU. A vLLM~\cite{kwon2023efficient} implementation reduces the evaluation time to 15 minutes. We do not discuss the efficiency aspect as we focus on the accuracy of the evaluator. Future work could build more time-efficient evaluators that are as accurate.
    \item The majority of data is GPT-4 generated. We do not have a guarantee of the data quality, but manually inspecting a small portion of the data indicates that it is not a serious issue.   
\end{itemize}

\section*{Acknowledgements}
We thank the members of UT NLP, especially Fangyuan Xu, for their feedback on the draft. The project is partially supported by a grant from Open Philanthropy and NSF grant RI-2312948. We also thank Chitrank Gupta for contributing to the earlier phase of this work. We also thank Ayon Das, an undergraduate student at UT Austin, to help with perspective detection annotation. This work was supported in part through the NYU IT High Performance Computing resources, services, and staff expertise.

\ifacl@anonymize
 
\else

\fi

\bibliography{custom,anthology}

\appendix

\newpage

\section{Experimental Details}
\label{sec:appendix}

\begin{table}
\footnotesize

\begin{center}
\begin{tabular}{lccc}
\toprule
Corpus &BM25 & DPR & \textsc{Contriever}  \\ \midrule
  Wikipedia & 0.99 & 0.95 & 0.75 \\
  Sphere & 0.99 & 0.99 & 0.95\\
  Google Search Output & 0.95 & - & 0.95\\
\bottomrule
\end{tabular}
\end{center}
\caption{The $\lambda$ value in~\Cref{eq:mmr} for different retriever-corpus setting.  }
\label{tab:lambda}
\end{table}

\begin{table*}
\fontsize{8}{10}\selectfont

\begin{center}
\begin{tabular}{p{1.5cm} | p{13cm}}
\toprule
Setting & Prompt \\ \midrule


Question Generation (2-shot) & Document 1: Natural habitats being are destroyed A tougher approach to the protection of animals is needed to prevent their natural habitats from being destroyed by locals. As humans expand their agricultural activity in Africa they are destroying the environments of endangered animals and pushing others towards being endangered. Due to an increase in large scale cotton plantations and food crops, the West African lion has seen a marked decrease in population; numbering less than 400 in early 2014 [1] . Tougher protection, such as fencing off areas from human activity, has been suggested and has seen success in South Africa [2] . [1] BBC, “Lions ‘facing extinction in West Africa’” [2] Morelle,R. “Fencing off wild lions from humans ‘could save them’”

Document 2: Fewer human deaths Fewer large beasts will lead to fewer deaths in Africa. Some endangered animals are aggressive and will attack humans. Hippopotamuses kill in excess of three hundred humans a year in Africa, with other animals such as the elephant and lion also causing many fatalities. [1] Footage released in early 2014 of a bull elephant attacking a tourist’s car in Kruger National Park, South Africa demonstrated the continued threat these animals cause. [2] Tougher protection would result in higher numbers of these animals which increases the risk to human lives. [1] Animal Danger ‘Most Dangerous Animals’ [2] Withnall, A. ‘Rampaging bull elephant flips over British tourist car in Kruger Park’

Instruction: Given the two documents, generate a question where both documents could be valid evidence. The question should be concise, and should not copy verbatim from the documents. It should be a Yes/No question where one document would support the "Yes" answer and the other would support "No".

Question: Should we encourage more intense protection of endangered species?  \\

& \\

 & Document 1: Allowing the sale of generic drugs will not help the plight of the developing world. Many drug companies invest substantial amounts of money, gleaned from the sale of profitable dugs in the developed world, into researching treatments for the developing world. Without the revenues available from patent-protected drug sales, companies' profits will fall, precipitating a reduction in pro bono giving and research. Allowing the production of generic drugs will thus in the long run hurt the developing world.

Document 2: Allowing production of generic drugs saves lives, particularly in the developing world Many developing countries are fraught with terrible disease. Much of Africa and Asia are devastated by malaria, and in many parts of Africa AIDS is a horrendous scourge, infecting large percentages of many countries populations. For example, in Swaziland, 26\% of the adult population is infected with the virus1. In light of these obscenely high infection rates, African governments have sought to find means of acquiring enough drugs to treat their ailing populations. The producers of the major AIDS medications do donate substantial amounts of drugs to stricken countries, yet at the same time they charge ruinously high prices for that which they do sell, leading to serious shortages in countries that cannot afford them. The denial of the right to produce or acquire generic drugs is effectively a death sentence to people in these countries. With generic drugs freely available on the market, the access to such drugs would be facilitated far more readily and cheaply; prices would be pushed down to market levels and African governments would be able to stand a chance of providing the requisite care to their people2. Under the current system attempts by governments to access generic drugs can be met by denials of free treatments, leading to even further suffering. There is no ethical justification to allow pharmaceutical companies to charge artificially high prices for drugs that save lives. Furthermore, many firms that develop and patent drugs do not share them, nor do they act upon them themselves due to their unprofitability. This has been the case with various treatments for malaria, which affects the developing world almost exclusively, thus limiting the market to customers with little money to pay for the drugs3. The result is patents and viable treatments sitting on shelves, effectively gathering dust within company records, when they could be used to save lives. But when there is no profit there is no production. Allowing the production of generic drugs is to allow justice to be done in the developing world, saving lives and ending human suffering. 1 United Nations. 2006. "Country Program Outline for Swaziland, 2006-2010". United Nations Development Program. Available: 2 Mercer, Illana. 2001. "Patent Wrongs". Mises Daily. Available: 3 Boseley, Sarah. 2006. "Rich Countries 'Blocking Cheap Drugs for Developing World'". The Guardian. Available:

Instruction: Given the two documents, generate a question where both documents could be valid evidence. The question should be concise, and should not copy verbatim from the documents. It should be a Yes/No question where one document would support the "Yes" answer and the other would support "No".

Question: Should we allow the production of generic drugs? \\

& \\

& Document 1: [Doc 1]

Document 2: [Doc 2]

Instruction: Given the two documents, generate a question where both documents could be valid evidence. The question should be concise, and should not copy verbatim from the documents. It should be a Yes/No question where one document would support the "Yes" answer and the other would support "No".

Question:  \\ \midrule 
Perspective Generation & Convert the question into a statement without adding any extra information. Then also generate the negation of this statement. 

Question: [Question] \\

\bottomrule
\end{tabular}
\end{center}
\caption{The prompt we use for generating data for Arguana. [Doc 1], [Doc 2], and [Question] are substituted with the actual documents and questions during generation. The question is first generated with the prompt in the first row, and then perspectives are generated with the prompt in the second row based on the question. }
\label{tab:prompts-arguana}
\end{table*}

\begin{table*}
\footnotesize

\begin{center}
\begin{tabular}{ l|p{14cm}}
\toprule
 Type & Prompt \\ \midrule


Yes/No &
Given the question and a positive statement, generate a statement of opposite stance with respect to the question.

Positive Statement: [Statement]

Question: [Question]

Negative Statement: \\ \midrule 
 Multi & Given the answers and the question, generate statements that are followed by a newline character. The statement should closely follow the format of the question, and the statement should only differ by the segment where the answer is.

Question: [Question]

Answers: [Answers]

Statements:
 \\

\bottomrule
\end{tabular}
\end{center}
\caption{The prompts we use for generating data for Kialo. Different prompts are used according to the data type (whether the question allows more than two perspectives). [Statement],[Question], and [Answers] are substituted with the statement, the question, and the answers provided by \url{Kialo.com} respectively during generation. Note that a list of answers is only provided for the questions with more than two perspectives. }
\label{tab:prompts-kialo}
\end{table*}

\begin{table*}
\vspace{-1em}
\fontsize{8}{10}\selectfont
\vspace{-1em}
\begin{center}
\begin{tabular}{p{2cm} p{6cm} p{6cm}}
\toprule
 Dataset & Original Data (Provided by the Source) & Generated Data \\ \midrule
\textbf{Arguana} & 
\hlc[gray]{\textbf{Document 1}}: 

People are given too much choice, which makes them less happy.  Advertising leads to many people being overwhelmed by the endless need to decide between competing demands on their attention - this is known as the tyranny of choice or choice overload. ...

\hlc[gray]{\textbf{Document 2}}: 

People are unhappy because they can't have everything, not because they are given too much choice and find it stressful. In fact, advertisements play a crucial role in ensuring ...
&
\hlc[lightblue]{\textbf{Question}}: 

Does an increase in choice lead to unhappiness and stress?

\hlc[lightred]{\textbf{Supporting Perspective}}: 

An increase in choice leads to unhappiness and stress.  

\hlc[lightred]{\textbf{Opposing Perspective}}: 

An increase in choice does not lead to unhappiness and stress. \\ \midrule
 
 \textbf{Kialo (Yes/No Question)} & 
\hlc[lightblue]{\textbf{Question}}: 

Will ChatGPT do more harm than good?

\hlc[lightred]{\textbf{Supporting Perspective}}: 

ChatGPT will do more harm than good.  &

\hlc[lightred]{\textbf{Opposing Perspective}}: 

ChatGPT will do more good than harm.  \\ \midrule
 \textbf{Kialo (More than two perspectives)} & 
\hlc[lightblue]{\textbf{Question}}: 

What's the best Twitter/X alternative?

\hlc[gray]{\textbf{Answer1}}: Facebook

\hlc[gray]{\textbf{Answer2}}: Kialo

\hlc[gray]{\textbf{Answer3}}: Instagram

\hlc[gray]{\textbf{Answer4}}: Reddit

& 
\hlc[lightred]{\textbf{Perspective 1}}: 

The best Twitter/X alternative is Facebook. 

\hlc[lightred]{\textbf{Perspective 2}}: 

The best Twitter/X alternative is Kialo. 

\hlc[lightred]{\textbf{Perspective 3}}: 

The best Twitter/X alternative is Instagram. 

\hlc[lightred]{\textbf{Perspective 4}}: 

The best Twitter/X alternative is Reddit.  

...  \\ \midrule
\textbf{OpinionQA} & 
\hlc[gray]{\textbf{Original Question}}:

Would having a gun in your household make you feel  ['Safer than you feel without a gun in your household', 'Less safe than you feel without a gun in your household', 'No more or less safe', 'Refused']

&
\hlc[lightblue]{\textbf{Generated Question}}: 

Do you feel safer with a gun in your household?

\hlc[lightred]{\textbf{Perspective 1}}: 

Having a gun in the household increases my sense of safety.

\hlc[lightred]{\textbf{Perspective 2}}: 

Having a gun in the household does not increase my sense of safety.  \\  

 \bottomrule
\end{tabular}
\end{center}

\vspace{-1em}
\caption{An example of data provided by the original source vs. generated. Data in the ``Original Data'' column is either provided by the source dataset (Arguana and OpinionQA) or \url{Kialo.com}. \url{Kialo.com} provide different information for different types of questions (whether the question is a yes/no question or not). Questions in \hlc[lightblue]{\textbf{blue}} and perspectives in \hlc[lightred]{\textbf{red}} are the ones we include in \bench{}. }
\label{tab:dataset-ex-compare}\vspace{-0.5em}
\end{table*}

\begin{table*}
\fontsize{8}{10}\selectfont

\begin{center}
\begin{tabular}{p{1.5cm}|p{13.5cm}}
\toprule
 Setting & Prompt \\ \midrule

 Question Filtering & Determine if the question is asking about personal experience or information. Reply with only "Yes" or "No".
Only answer "Yes" if 
(1) The question is asking about if you have ever done something or if something has happened to you.
(2) The question is asking about facts about you, for example, if you own something, or how you describe yourself.
(3) The question is asking about something that has happened to your local community, or the current status of your local community.
You should answer "No" if the question is asking about your opinion, such as if you are worried about something or what you think about a certain subject. 

Question: [Question] 

Answer: 
 \\ \midrule
 
Data Generation &
Convert the following survey question into a Yes/No question, with the first and the last choice in the brackets as the two answers. Generate the question first, and then convert the question into a positive and a negative statement that supports and oppose the question respectively. The statements should be concise and not contain any additional information that is not in the question. The statements should also be fluent and grammatically correct.

Question: [Question] 

Converted question: \\

\bottomrule
\end{tabular}
\end{center}
\caption{The prompt we use for generating data for OpinionQA. [Question] is substituted with the original question, which comes with a list of options, in the OpinionQA dataset during generation. We first filter out data that are not related to personal experience or facts about the survey taker. Then we generate questions and perspectives based on the original question provided by OpinionQA. }
\label{tab:prompts-opinionqa}
\end{table*}

\begin{table*}
\fontsize{8}{10}\selectfont

\begin{center}
\begin{tabular}{p{15cm}}
\toprule
Generate diverse perspectives that answer the given question. Please answer with a JSON object, where each field is a perspective. Do not generate similar perspectives; they should be distinct. The keys in the JSON object should be meaningful, and each word in the keys should be separated by whitespace. 

Question: [Question] \\
\bottomrule
\end{tabular}
\end{center}
\caption{The input prompt for generating diverse perspectives for query expansion. [Question] is substituted with the question of each example. The output would be a JSON object containing the perspectives. }
\label{tab:prompts-div}
\end{table*}

\begin{table*}
\fontsize{7.7}{10}\selectfont
\begin{center}
\begin{tabular}{p{1.5cm} p{1.8cm} p{11.5cm}}
\toprule
 Dataset & Question & GPT-4-Generated Perpsectives \\ \midrule
\textbf{Arguana} & 
\hlc[lightblue]{\textbf{Question}}: 

Does an increase in choice lead to unhappiness and stress?
&

\hlc[lightred]{\textbf{Increased\_Choice\_Leads\_to\_Paralysis}}: An increase in choice may lead to feelings of unhappiness and stress, due to the decision paralysis phenomena. With too many options, individuals may struggle to make a choice and fear making the wrong decision.

\hlc[lightred]{\textbf{Choice\_Overload\_Stress}}: When faced with numerous choices, individuals may feel overwhelmed, raising their stress levels. Too many options may make decision-making processes lengthy and taxing, escalating stress and leading to unhappiness.

\hlc[lightred]{\textbf{Regret\_Fear\_and\_Happiness\_Decrease}}: Increased choice can potentially lead to decreased happiness and increased stress, as individuals may worry about their decisions, regret not choosing different options, or fear missing out on potentially better choices.
 \\ \midrule
 
 \textbf{Kialo} & 
\hlc[lightblue]{\textbf{Question}}: 

Will ChatGPT do more harm than good?

  &

\hlc[lightred]{\textbf{Artificial\_Intelligence\_Opportunity}}: ChatGPT, like other AI models, can have immense potential and bring about substantial benefits. By enabling natural language interaction, it can be used in areas such as education, healthcare, customer service, and more, greatly enhancing efficiency and user experience. The potential good that ChatGPT can contribute far outweighs the potential harm, assuming that the proper controls and ethical guidelines are put in place.

\hlc[lightred]{\textbf{Miscommunication\_Risk}}: While ChatGPT is designed to understand and generate human-like text, there's a risk of miscommunication. The AI model may not fully grasp subtle nuances, cultural contexts, or emotions which could lead to misunderstanding or even harm, especially in sensitive conversation scenarios.

\hlc[lightred]{\textbf{Data\_Privacy\_Concern}}: There's a potential for misuse pertaining to data privacy with ChatGPT. It's trained on incredibly large datasets, and there might be instances where personal or sensitive data might be leaked or misused, posing a serious threat to privacy.
\\ \midrule

\textbf{OpinionQA} & 

\hlc[lightblue]{\textbf{Question}}: 

Do you feel safer with a gun in your household?

&

\hlc[lightred]{\textbf{Second\_amendment\_advocate\_perspective}}: 
Yes, I do feel safer with a gun in the household. It provides a means of defending my family, property and myself from potential offenders or burglars.

\hlc[lightred]{\textbf{Child\_safety\_concerned\_perspective}}: 
No, I don't feel safer with a gun in the home. Especially when there are kids around, the risk of accidents due to negligence is significant and unnerving. 

\hlc[lightred]{\textbf{Non\_violence\_promoter\_perspective}}: No, owning a gun insinuates the readiness for violence, which creates an atmosphere of constant threat and insecurity. I believe in resolving conflicts peacefully rather than aggressively.
\\  

 \bottomrule
\end{tabular}
\end{center}

\vspace{-1em}
\caption{Examples of diverse perspectives generated by GPT-4, later used for query expansion in~\Cref{ssec:div-aware}. We only present the first three generated perspectives due to the page limit.}
\label{tab:gpt-pers-ex}
\end{table*}

\subsection{Prompts and Examples for Data Generation}
\label{ssec:prompt-data}
Prompts for generating each dataset are shown in~\Cref{tab:prompts-arguana,tab:prompts-kialo,tab:prompts-opinionqa}. Examples showing the generated data and data taken from the source can be found in~\Cref{tab:dataset-ex-compare}.

\subsection{Prompts for Perspective Detection}
\label{ssec:prompt-pers}
Prompts we use for perspective detection are shown in~\Cref{tab:prompts-pers-det}. We prompt all models with the prompt in the top row in~\Cref{tab:prompts-pers-det}, and additionally report results prompting GPT-4 with the in-context example shown in the bottom row (the last row and the best performance in~\Cref{tab:pers-det}). The labels of the data used to fine-tune Mistral-7B are predicted with the one-shot prompt.   

\begin{table*}
\fontsize{8}{10}\selectfont

\begin{center}
\begin{tabular}{p{1.5cm}|p{13.5cm}}
\toprule

Prompt & <|im\_start|>system

You are a helpful assistant that can only answer with "Yes" or "No". Your goal is to determine if the provided document support the provided statement. Do NOT provide any explanation for your choice.

<|im\_end|>

[Optional 1-shot example]

<|im\_start|>user

Document: [Document]

Statement: [Statement]

Instruction: Does the document support the statement? Answer with only "Yes" or "No". 
Only consider the information provided in the document, and do not infer any additional information. 
The document is a snippet taken from the Internet, and might exhibit support for or opposition against the statement.
If the document opposes the statement, the answer is "No". If the document does not contain relevant information about the statement, the answer is also "No".
The answer is "Yes" only if the document supports the statement, either explicitly or implicitly.
Think carefully if the document actually supports the statement, or if there is just some superficial textual overlap between the two.

Answer:

<|im\_end|> \\ \midrule
One-Shot Example & <|im\_start|>user

Document for Example 1: Why Prop. 19 is so important | rescue truth. The United States took a stab at alcohol prohibition from 1920-1933 via the United States Constitution. Almost immediately following the ratification of the Eighteenth Amendment, which outlawed alcohol, speakeasy clubs sprung up all over, effectively creating a black market and handing alcohol sales over to organized crime. During this time, our government poisoned industrial alcohol in an attempt to curb usage by scaring the American public, causing an estimated 10,000 deaths.[17] Because of the illicit drug market created by prohibition, the government could not regulate the production of alcohol, and sometimes people became ill after drinking bootlegged whiskies. The current

Statement for Example 1: Prohibition is an effective method to curb drug usage.

Instruction for Example 1: Does the document support the statement? Answer with only "Yes" or "No". 

Only consider the information provided in the document, and do not infer any additional information. 
The document is a snippet taken from the Internet, and might exhibit support for or opposition against the statement.
If the document opposes the statement, the answer is "No". If the document does not contain relevant information about the statement, the answer is also "No".
The answer is "Yes" only if the document supports the statement, either explicitly or implicitly.
Think carefully if the document actually supports the statement, or if there is just some superficial textual overlap between the two.

Answer for Example 1:

<|im\_end|>

<|im\_start|>assistant

No

<|im\_end|> \\

\bottomrule
\end{tabular}
\end{center}
\caption{The input prompt for perspective detection. We provide an optional one-shot example for some models that do not do well zero-shot. [Document] and [Statement] are substituted with the document and perspective in each example respectively.  }
\label{tab:prompts-pers-det}
\end{table*}

\begin{table*}
\fontsize{8}{10}\selectfont

\begin{center}
\begin{tabular}{p{15cm}}
\toprule
Question: [Question]

Perspective 1: [P1]

Perspective 2: [P2]

Instruction: Given the question, one supporting and one opposing perspectives, which perspective is supporting the question? Answer with only ``Perspective 1'' or ``Perspective 2''.

Answer:
\\
\bottomrule
\end{tabular}
\end{center}
\caption{The input prompt for deciding which of the perspectives supports or opposes the question. We only consider examples with two perspectives. We replace [Question] with the actual question, and [P1], [P2] with the two perspectives when prompting. }
\label{tab:prompts-pos-op}
\end{table*}

\subsection{Prompts for Query-Expansion}
\label{ssec:prompt-div}
Prompts we use to generate perspectives for query-expansion are shown in~\Cref{tab:prompts-div}.


\subsection{Details on Constructing Perspective Detection Test Set}
\label{ssec:pers-det-test}
Each example in the test set consists of a perspective and a document. We randomly select 10 questions with two perspectives from each of the three datasets, yielding a total of 60 perspectives. For each question, we take the top five retrieved documents from either BM25 or DPR results obtained from Sphere. Thus there are ten documents to test for each perspective, resulting in 600 document-perspective pairs. We then check for near-duplicate documents and filter out 58 examples, leaving us with 542 examples.

\subsection{Details of Fine-tuning Mistral-7B Models}
\label{ssec:finetune}

We fine-tune the Mistral-7B model with LoRA~\cite{hu2021lora} on eight NVIDIA A40 GPUs. We train the model on 3K examples of GPT-4 labeled data on perspective detection. The training dataset is constructed using the same process as described in~\Cref{ssec:pers-det-test}, except on 50 questions, yielding five times more data. 

The model is trained using an AdamW~\cite{loshchilov2018decoupled} optimizer with a linear learning schedule. We perform a hyperparameter search on learning rates (1e-4 2e-4 1e-5 2e-5 3e-5 5e-5), warmup ratio (0.1 0.3 0.5), number of training epochs (0.5, 1, 2, 3, 4), and weight decay (1e-4, 0.001, 0.01, 0). We evaluate each checkpoint on a separate development set of 100 examples annotated by the author and select the model with the highest F1 score. The final hyperparameters we decided on are learning of 1e-4, a warmup ratio of 0.1, 2 epochs of training, and a weight decay of 0.001. 

Since the training examples are constructed using the same retriever-corpus settings, we also report results on out-of-domain examples in \Cref{ssec:add-pers-det}. 

\subsection{Detail of Re-ranking Experiments}
\label{ssec:rerank-details}
We compute the $Sim_2$ scores in~\Cref{eq:mmr} by taking the cosine similarity between two document embeddings computed by a sentence embedding model, the UAE-Large-V1 model on huggingface~\footnote{\url{https://huggingface.co/WhereIsAI/UAE-Large-V1}}. We follow their documentation for non-retrieval tasks and set the \texttt{pooling-strategy} =\texttt{`cls'}. We do not specify any prompt for the model so the embeddings are only on the documents themselves. 

$Sim_1$ are obtained from the implemented retriever model, and thus they could be of different scales. We then normalize the $Sim_1$ scores to be between [0, 1] by dividing $Sim_1$ by the maximum score obtained on the whole dataset, and do a grid search on $\lambda$ over values [0.5, 0.75, 0.9, 0.95, 0.99]. The actual values we choose for each retriever-corpus setting are shown in~\Cref{tab:lambda}.

\subsection{Examples of Generated Perspectives for Query Expansion}
\label{ssec:ex-diversity-aware}
We present examples of perspectives generated by GPT-4 on each dataset in~\Cref{tab:gpt-pers-ex}. Each perspective has a name separated by ``\_", and the corresponding text. The average length of perspectives is 43.98 words. 

\subsection{Details of Determining Supporting and Opposing Perspectives}
\label{ssec:details-sup-op}
We identify the supporting and opposing perspectives by prompting \texttt{gpt4-0613}, and the exact prompt can be found in~\Cref{tab:prompts-pos-op}. We only perform such a process for examples with two perspectives.

\begin{table*}
\footnotesize
\begin{center}

\begin{tabular}{ll cc cc cc |cc}
\toprule
 \multirow{3}{*}{Corpus} &\multirow{3}{*}{Model}  & \multicolumn{6}{c}{Datasets} \\
&&\multicolumn{2}{c}{Arguana} & \multicolumn{2}{c}{Kialo} & \multicolumn{2}{c}{OpinionQA} & \multicolumn{2}{c}{Average}  \\
 &   & Prec. & MRec.  & Prec. & MRec.  & Prec. 
 & MRec.& Prec. & MRec.  \\ \midrule

 \multirow{3}{*}{Wiki} & BM25&19.81&19.33&20.58&16.67&9.13&5.67&16.51&13.89 \\
 & DPR&17.35&14.67&18.17&13.57&12.03&6.35&15.85&11.53 \\
 & \textsc{Contriever}&32.37&31.07&30.71&26.10&21.50&13.83&28.19&23.67 \\ 
 &\textsc{Tart} & 34.53&35.47&31.53&26.53&23.40&15.07&29.82&	25.69 \\
 & NV-Embed-v2 & 36.87 & 36.67 & 35.89 & 32.13 & 26.52 & 14.53  & 33.09 & 27.78 \\
\midrule
 \multirow{3}{*}{Sphere} & BM25&42.44&43.20&41.54&37.34&44.09&33.67&42.69&38.07\\
 & DPR&44.09&33.67&13.51&12.40&8.16&6.12&21.92&17.40 \\
 & 
\textsc{Contriever}& 55.89  & 55.07  & 53.89  & 43.28  & 48.61  & 38.66  & 52.80  &45.67 \\
& \textsc{Tart} & 65.44&59.87&56.45&44.80&\textbf{63.08}&41.33&61.66&48.67 \\
& NV-Embed-v2 & \textbf{68.03}&\textbf{64.00}&\textbf{64.74}&\textbf{54.39}&60.63&\textbf{48.07}&\textbf{64.47}&\textbf{55.49} \\
\midrule
 & BM25&33.0&36.40&34.92&38.50&33.81&29.25&33.91&34.72 \\
  Google Search & \textsc{Contriever}&48.65&44.53&47.95&44.32&42.73&35.49&46.44&41.45 \\
 Output& \textsc{Tart} & 51.43&45.33&49.99&43.20&45.69&35.07&49.04&41.20 \\
 & NV-Embed-v2 & 59.17& 55.87&61.61&51.60&50.32&40.13&57.03&49.20 \\
\bottomrule
\end{tabular}
\end{center}
\caption{Performance on the test split of \bench{}, considering top 10 retrieved documents.}
\label{tab:k10-results}
\end{table*}

\newpage
\section{Additional Results}
\label{sec:add-results}

\subsection{Results for Top 10 Documents}
\label{ssec:results-k10}
We report results for retrieving $k$ = 10 documents in~\Cref{tab:k10-results}. 
Comparing the results with those in~\Cref{tab:main-results} shows that including more documents leads to higher diversity (\textsc{MRecall}) at the cost of \textsc{Precision} due to the reduced relevance of lower-ranked documents. Similar trends hold for $k$ = 10, as \textsc{Contriever} is still the best retriever in terms of diversity, and retrievers obtain more diverse documents on Sphere, followed by Google Search API outputs and Wikipedia.

\begin{table}
\footnotesize
\begin{center}
\vspace{-0.5em}
\begin{tabular}{ll rr}
\toprule

{Corpus} & {Model} & Avg. size ($\downarrow$) & \% within top 100 ($\uparrow$)\\
\midrule
 

\multirow{3}{*}{Wiki} & BM25 &29.8&41.5\\
 & DPR &30.9&36.7\\
& \textsc{Cont.} &24.7&55.6\\ \midrule 
 \multirow{3}{*}{Sphere} & BM25 &20.8&77.3\\
& DPR &37.4&48.9\\
 & \textsc{Cont.} &17.8&83.1\\ \midrule
 \multirow{2}{*}{Google} & BM25 &21.0&72.1\\
& \textsc{Cont.} &18.2&76.2\\
\bottomrule
\end{tabular}
\end{center}\vspace{-0.5em}
\caption{The average size of the document set required to cover all perspectives with base retrievers. The average size is only computed over the examples where the sets of the top 100 documents contain all perspectives (\textsc{MRecall} @ $100$ = 1), and we report \% of data that satisfies this (\textsc{MRecall} @ $100$) in the last column.}
\label{tab:rank}
\vspace{-0.5em}
\end{table}

\subsection{How many documents should one retrieve to cover diverse perspectives?} \label{ssec:rank}
In~\Cref{tab:main-results}, we have reported the performances of the retrieval outputs consisting of five documents and found that the model outputs do not cover diverse perspectives. If we consider up to the top 100 documents, can we cover diverse perspectives? What is the average number of documents needed to be retrieved to cover all perspectives? Table~\ref{tab:rank} reports the size of the retrieved document set needed to cover all the perspectives with base retrievers. We find that the best setting (\textsc{Contriver} with Sphere) can find all the perspectives within the top 100 document set 83.1\% of the time, and for those examples, the average number of documents required to cover all the perspectives is 17.8. Other settings show worse performance, mostly reflecting performance reported in Table~\ref{tab:main-results}.

\subsection{Additional Validation Results on Perspective Detection}
\label{ssec:add-pers-det}
The training examples for fine-tuning the perspective detection model share the (corpus, retriever) settings for evaluation. We thus consider the results in~\Cref{tab:pers-det} to be in-domain. 
We conduct further evaluation on out-of-domain examples. We sample 300 (Perspective, Document) pairs per (corpus, retriever) setting. We label each pair with GPT-4, and compare the trained Mistral perspective detection model against GPT-4 predictions. The results are in~\Cref{tab:add-pers-det}. The Mistral model is trained with GPT-4 labeled examples from the bolded settings (Sphere+BM25, Sphere+DPR). Our results show that it generalizes to out-of-domain settings.

\begin{table}
\footnotesize
\begin{center}
\vspace{-0.5em}
\begin{tabular}{llrrr}
\toprule

Corpus	&Retriever&	Acc.&	F1&	\% Positive  \\
&&&& (Labeled  \\
&&&& by GPT-4) \\ \midrule

Wiki &	BM25	&92.3&	68.1	&12.7 \\
Wiki &	DPR	&92.1	&70.4	&14.2 \\ 
Wiki &	Contriever	&91.1	&71.6	&16.3 \\ \midrule
\textbf{Sphere} &	\textbf{BM25}	&\textbf{94.0}	&\textbf{83.0}	&\textbf{13.3} \\ 
\textbf{Sphere} &	\textbf{DPR}&	\textbf{97.0}	&\textbf{60.9}	&\textbf{12.7} \\ 
Sphere & Contriever&	89.3	&75.8	&14.2 \\ \midrule
Google Search 	& \multirow{2}{*}{BM25}	&\multirow{2}{*}{91.5}	&\multirow{2}{*}{76.7}	&\multirow{2}{*}{19.2} \\
Output & &&& \\
Google Search	&\multirow{2}{*}{Contriever}&	\multirow{2}{*}{89.0}	&\multirow{2}{*}{76.3}	&\multirow{2}{*}{24.3} \\
Output & &&& \\

\bottomrule
\end{tabular}
\end{center}\vspace{-0.5em}
\caption{Out-of-domain evaluation for the trained perspective detection model. The settings used to train the perspective detection model are \textbf{bolded}. The out-of-domain performances are roughly comparable to in-domain performances. }
\label{tab:add-pers-det}
\vspace{-0.5em}
\end{table}

\subsection{Manual Quality Inspection of BERDS}
\label{ssec:manual-errors}
We manually examine 20 random examples in each subset of BERDS. The number of errors in each subset is presented in~\Cref{tab:manual-error-rate}. We find that only one example in Arguana contains an unfaithful generated question, and the question itself is a valid contentious question (\Cref{tab:manual-error-arguana}). In other words, the minor error does not affect the validity of the generated data. Only one example has a minor error (see~\Cref{tab:manual-error-kialo}) in Kialo, and it is because the provided positive perspective has a slight misalignment with the provided question. None of the examples in OpinionQA contain errors. The errors are minor and one can even argue the generated data is valid. Thus, we can infer that the generated data are of high quality.

\begin{table}
\footnotesize
\begin{center}
\vspace{-0.5em}
\begin{tabular}{rrr}
\toprule

{Arguana} & Kialo & OpinionQA\\
\midrule
  1 & 1 & 0  \\
\bottomrule
\end{tabular}
\end{center}\vspace{-0.5em}
\caption{We manually examine 20 random examples of each subset in BERDS, and show the number of errors. Only one example contains a minor error in each of Arguana and Kialo.}
\label{tab:manual-error-rate}
\vspace{-0.5em}
\end{table}

\begin{table*}
\footnotesize
\begin{center}
\vspace{-0.5em}
\begin{tabular}{p{1.5cm} p{2cm} p{6.3cm} p{4.2cm}}
\toprule

Question (Generated)&Perspectives (Generated)	&Document 1 (Provided)&Document 2 (Provided) \\ \midrule
Is unrestricted access to birth control beneficial for society?&	['Unrestricted access to birth control is beneficial for society.', 'Unrestricted access to birth control is not beneficial for society.']&	Any body of values that claims to respect the rights of the individual must recognise the right of a woman to choose Even the doctrines of the Church accepts that pregnancy is not, in and of itself, a virtue – there is no compulsion to maximise the number of pregnancies; there is simply a disagreement about how they should be avoided. The Church recommends that couples may minimise the chance without ever making it impossible through a chemical or physical barrier. In some parts of the world a pregnancy, even one that is not planned, is seen as a time for joy – a blessing for the family that will lead to a new and happy life bringing pleasure to both parents, their society and the child. That ideal is very far from the experience of much of the world where a child is another mouth to feed on impossibly little income. For all too much of the world, that life will be cruel, nasty and short. In slums, favellas and barren wastes that life is likely to be one marked more by dysentery or diarrhea, malnutrition and misery than by the sanitised, idealised image promoted in the West. [...]	&It is difficult to see how the life of anyone is improved by reducing sex to a cheap form of entertainment. Certainly not the unborn children and not the objectified women. Proposition is more than happy for women to take control of their own fertility – indeed we would go further and suggest that their boyfriends and husbands should do so as well. Recreational sex, within wedlock and during times of infertility removes all of these problems; a little planning and restraint achieves that aim. It also means that both parents need to show that they are responsible for the results; Op seems happy to say that people are uncontrollable beasts with no control over their desires – hardly an edifying concept. \\

\bottomrule
\end{tabular}
\end{center}\vspace{-0.5em}
\caption{Error Case of Arguana. The generated question may not have grasped the nuance of the discussion of the two provided documents, but it is relevant to the subject (birth control) and is a valid contentious question. }
\label{tab:manual-error-arguana}
\vspace{-0.5em}
\end{table*}

\begin{table*}
\footnotesize
\begin{center}
\vspace{-0.5em}
\begin{tabular}{p{3.8cm} p{5.2cm} p{5.2cm}}
\toprule

Question (Provided) & 	Positive Perspective (Provided)	 & Negative Perspective (Generated) \\		\midrule
Should There be a Universal Basic Income (UBI)?	&Wealthy countries should provide citizens with a universal basic income (UBI).&	Wealthy countries should not provide citizens with a universal basic income (UBI). \\

\bottomrule
\end{tabular}
\end{center}\vspace{-0.5em}
\caption{The error case of Kialo. The generated negative perspective is not faithful to the provided question only because the provided positive perspective itself mentions ``wealthy countries". }
\label{tab:manual-error-kialo}
\vspace{-0.5em}
\end{table*}

\end{document}